# Multidimensional Service Quality Scoring System

Shiyang Lai [a, 1]

*a Division of Social Sciences, University of Chicago, Chicago 60637, USA*

This supplementary material aims to introduce the Multidimensional Service Quality Scoring System (MSQs), a review-based method for quantifying host service quality mentioned and employed in the paper *Exit and transition: Exploring the survival status of Airbnb listings in a time of professionalization*. The framework presented in Figure 1 summarizes the workflow of MSQs. MSQs is not an end-to-end implementation and is essentially composed of three pipelines, namely Data Collection and Preprocessing, Objects Recognition and Grouping, and Aspect-based Service Scoring. Using the study mentioned above as a case, the details of MSQs are explained below.

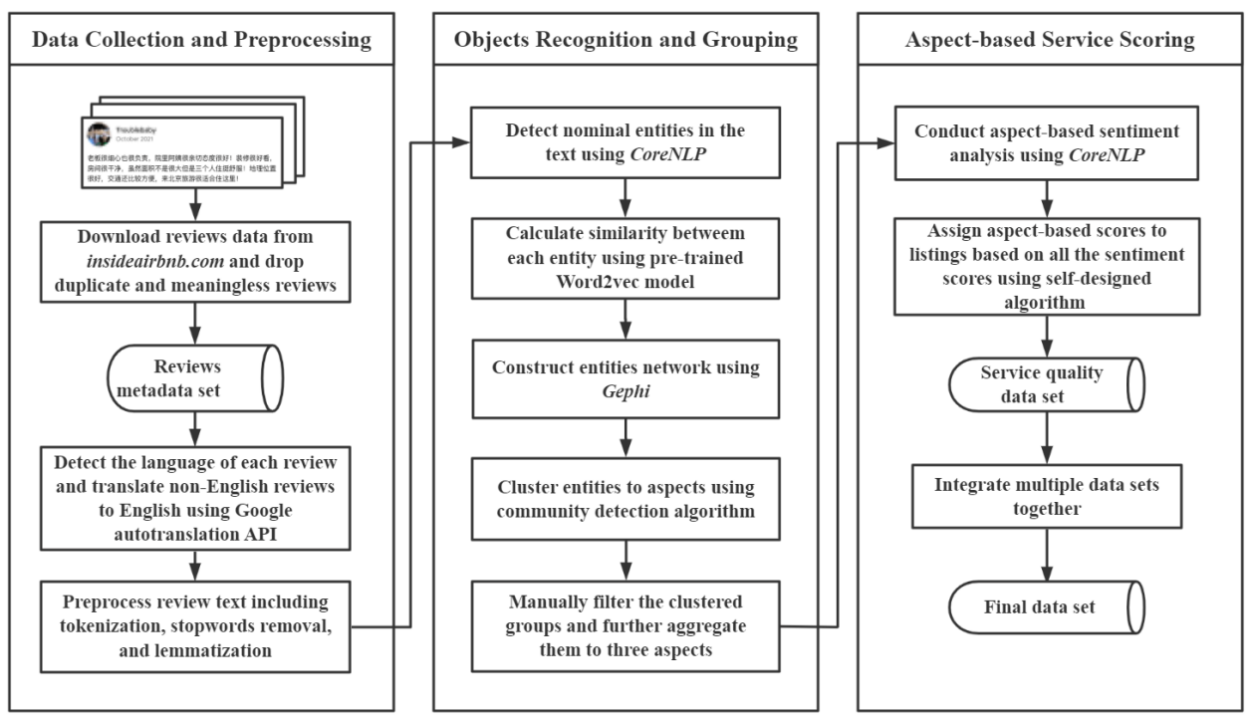

**Fig. 1.** The working framework of MSQs in the case of Airbnb

*Data Collection and Preprocessing*

This pipeline can be further divided into three substeps. The very first step is acquiring metadata. Experiment data should be organized in listing-review format and invalid reviews, such as those that only contain meaningless special characters, need to be cleaned. In our case, we obtained data from Airbnb in Beijing from July 2018 to June 2020, with 483,075 reviews in total. The next step is language standardization, but only for a multi-linguistic corpus. Since our review set is a mix of English and Chinese, we used the Google Translation interface to translate non-English reviews into English to unify language for subsequent analysis. The last step is review text preprocessing. Tokenization, stopwords removal, and lemmatization should be performed sequentially to further normalize text to a clean format.

[1]Corresponding author.
*E-mail addresses*: shiyanglai@uchicago.edu (S. Lai)

*Objects Recognition and Grouping*

The functions of this pipeline are to identify the primary aspects from reviews and extract the seed words for each aspect. Unlike existing methods, MSQs narrows the range of seed words through named entity recognition (NER) before extraction of aspects. Exploiting the toolkit provided by *CoreNLP*, we were able to find mentions of key "things" (i.e., entities) such as property attributes, hosts, and nearby facilities within the text and only treat these entities as potential seed words. Then, a pre-trained word embedding model should be employed to calculate the similarities between each entity pair. In our case, we employed a Word2Vec model pre-trained on the *glove-twitter-100* public corpus. The reason for choosing tweets data as a training data set instead of other popular or even larger corpuses like *Wiki* or *Google News* is that we believe that among all these alternatives the semantic context of tweets is closest to that of Airbnb online reviews.

By treating entities as nodes and the similarities between them as edges, a words network can be built. Ideally, this network contains all the words with actual meaning in the review dataset. Based on this, we can then cluster entities into groups through community detection algorithms. Community detection is a very important branch of network analysis theory, and this technique enables us to find multiple clusters with nodes inside them that are densely connected. For words networks, these clusters represent sets of words with similar linguistic contexts. Our MSQs employed a fast large-scale network community detection algorithm proposed by Blondel, Guillaume, Lambiotte, and Lefebvre (2008). Since this algorithm cannot operate on fully connected networks, edges with degree (i.e., the similarity between two words) lower than 0.5 are filtered out. After manually deleting some groups that were too small and some groups that had no specific meaning, we finally obtained 18 clusters of words associated with service quality (see Table 1). The schematic visualization of word clustering is presented in Figure 2.

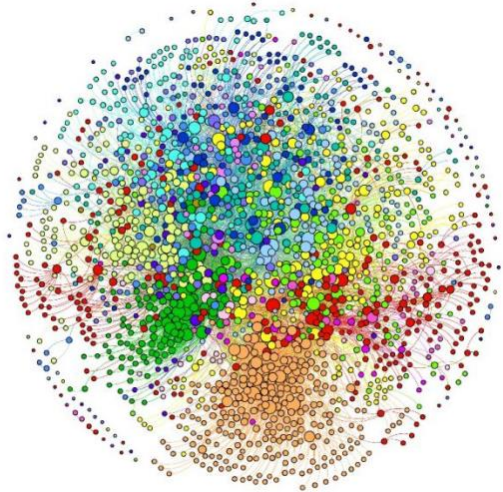

**Fig.2.** Words network and grouping result visualization

**Table 1**. Topic labeling

| No. | Topic Label | Representative Topical Words | Reference |
|---|---|---|---|
| 1. | Booking process | Check, checkout, booking, payment, app, detail, admission | Ding, Choo, Ng, and Ng (2020) |
| 2. | Cleanness | Clean, trash, neatness, tidying, sanitation, washing | Lawani, Reed, Mark, and Zheng (2019) |
| 3. | Climate | Weather, sunshine, cloud, storm, season, climate | Named by authors |
| 4. | Communication | Talk, advice, describe, telling, guide, tip, communication, chat, email | Luo and Tang (2019) |
| 5. | Event | Anniversary, party, entertainment, carnival, concert, festival | Named by authors |
| 6. | Experience | Feeling, experience, memory, mentality, sense | Luo and Tang (2019) |
| 7. | Facility | Freezer, microwave, bathtub, conditioner, elevator, television, computer, wireless, Wi-Fi, internet, hotspot | Cheng and Jin (2019) |
| 8. | Food and drink | Food, meal, eat, dinner, coffee, cook, tea | Lawani et al. (2019) |
| 9. | Host | Care, compliment, supporting, share, help, assistance, service, operation, attitude, skill | Cheng and Jin (2019) |
| 10. | Macro environment | Public, nature, citizen, country, area, community, local, surrounding, culture | Named by authors |
| 11. | Mate | Roommate, stranger, friend, flatmate, guest, homie | Named by authors |
| 12. | Nearby | Neighborhood, restaurant, barbershop, park, theater, bar, bakery | Lawani et al. (2019) |
| 13. | Property attribute | Condo, apartment, loft, penthouse, door, bedroom, lobby, space, architecture, renovation, decoration | Ding et al. (2020) |
| 14. | Security | Harm, scare, attack, guard, injury, risk, fear, surveillance | Named by authors |
| 15. | Shopping | Store, purchase, product, discount, shop, grocery | Ding et al. (2020) |
| 16. | Sleeping | Sleep, nightmare, nap, insomnia, bedding, wake, noise | Zhang (2019) |
| 17. | Transportation | Transport, taxi, metro, taxi, bus, railway, station, access, congestion, distance, route | Zhang (2019) |
| 18. | Value | Cost, value, worth, benefit, expense, fee | Luo and Tang (2019) |

*Note:* "Reference" suggests the source of topic name.

The last step of this pipeline is group aggregation, and it should be customized based on the specific interest of implementation. In terms of our study, because we wanted to investigate services with different levels of adjustability, we aggregated the 18 groups into 3 more abstract dimensions, namely: services that can be changed by the host with no additional costs, services that can be manipulated by the host but require additional costs, and services that cannot really be controlled by the host (see Table 2).

**Table 2**. Mapping of topic to service dimension

| Dimension | Description | Topic |
|---|---|---|
| High adjustability | Services that can be changed by the host with no additional costs | Communication, Mate, Cleanliness, Booking process, Host, Experience |
| Medium adjustability | Services that can be manipulated by the host but require additional costs | Facility, Property attribute, Food and drink, Security, Sleeping, Value |
| Low adjustability | Services that cannot really be controlled by the host | Climate, Event, Macro environment, Nearby, Shopping, Transportation |

*Aspect-based Service Scoring*

The last pipeline of MSQs rates the service quality provided by listing owners based on the aspects and the corresponding seed words (entities). Before jumping into tahe scoring logic, some preparation work needs to be done first. To measure guests' attitudes towards different aspects of services, the entities in each review are labeled using a sentiment processor provided by *CoreNLP*. If one sentence is positive, then all entities in it will be assigned the value 1. On the contrary, if the sentence is negative, all entities in it will be valued at -1. The next step is entity rescoring. The idea is that the sentiment of a sentence is the sum of the attitudes of the reviewer towards each entity mentioned in that sentence. We use $\alpha_{c,k,i,j}^{m}$ to represent the $j$th entity belonging to aspect $m$ in sentence $i$ of review $k$ of listing $c$. Mathematically, the rescoring strategy can be expressed as:

$$\alpha_{c,k,i,j}^{m}{}' = \frac{\alpha_{c,k,i,j}^{m}}{\sum_{j}^{J_i} \alpha_{c,k,i,j}^{m}},$$ (1)

where $J_i$ represents the number of entities appearing in sentence $i$.

Next, entity-level scores should be aggregated to review-level. For review $k$, by grouping and summing the entities appearing in $k$ according to the previously obtained aspects, the aspect-based service quality score of review $k$ can be obtained. The following formula illustrates the rule of aggregation:

$$\alpha_{c,k}^{m} = \sum_{i}^{I_k} \sum_{j}^{J_i} \alpha_{c,k,i,j}^{m}{}',$$ (2)

where $I_k$ is the number of sentences in review $k$.

Eventually, the overall score of service quality in different aspects can be found by calculating the mean of all reviews for one specific listing, that is

$$\alpha_c^m = \sum_k^{K_c} \alpha_{c,k}^m \,, \tag{3}$$

where $K_c$ is the number of reviews that listing $c$ has received.

*Merits of MSQs*

MSQs is not the first method for exploiting guests' reviews to rate hosts' service. Before it, Luo and Tang (2019) applied a modified latent aspect rating analysis (LARA) to extract hidden aspects in guests' reviews to assess how customers rated their homestay experiences. However, one essential problem related to their methodology is the Latent Dirichlet Allocation (LDA)-based aspect identification. Although LDA is still the most popular topic modeling method in academia, many researchers have asserted that it is not an appropriate tool when mining short texts (Hong & Davison, 2010; Qiang, Qian, Li, Yuan, & Wu, 2020). This is because relying entirely on experiment text that has extremely limited word co-occurrence information makes LDA incapable of generating confident estimations of the distribution of each topic. To solve this problem, we developed MSQs. Compared with LDA-based methods, MSQs tends to be less sensitive to the quality and quantity of experiment data since its word-embedding component is pre-trained on large data sets. Thus, MSQs has already gained sufficient knowledge to perform words classification before working on experiment text. Besides, following the belief of linguistic researchers that language is a network-like knowledge system, MSQs treats the semantic contexts of consumers' reviews as networks and utilizes community detection algorithms to do the topic extraction.

*Limitation of MSQs*

Although theoretically MSQs has some strengths that make it a potentially more suitable short-review-based service-quality-quantifying method than other, LDA-based methods, there are two main problems related to it. First, MSQs is not an end-to-end model, which means users need to implement each pipeline separately and some pipelines require manual operation. Thus, the implementation of MSQs is relatively harder and, to a certain extent, requires deeper understanding of the text itself. Second, MSQs is not expected to perform well when facing special linguistic contexts. In the *Objects Recognition and Grouping* pipeline, the pre-trained word-embedding model is usually based on corpuses generated from general linguistic contexts where the meaning of a word is often the top few definitions in a dictionary. But in some specific scenarios, the meanings of words change dramatically, differing significantly from ordinary usage. For instance, Arabic numbers and certain names refer to specific police codes, having very special meanings in the routine work of the police. In this scenario, MSQs will still understand numbers and names according to their most common meanings, thus causing misunderstanding and bias.

# References

Blondel, V. D., Guillaume, J.-L., Lambiotte, R., & Lefebvre, E. (2008). Fast unfolding of communities in large networks. *Journal of Statistical Mechanics: Theory and Experiment, 2008*, P10008.

Cheng, M., & Jin, X. (2019). What do Airbnb users care about? An analysis of online review comments. *International Journal of Hospitality Management, 76*, 58-70.

Ding, K., Choo, W. C., Ng, K. Y., & Ng, S. I. (2020). Employing structural topic modelling to explore perceived service quality attributes in Airbnb accommodation. *International Journal of Hospitality Management, 91*, 102676.

Hong, L., & Davison, B. D. (2010). Empirical study of topic modeling in Twitter. In *Proceedings of the First Workshop on Social Media Analytics* (pp. 80–88). Washington D.C., District of Columbia: Association for Computing Machinery.

Lawani, A., Reed, M. R., Mark, T., & Zheng, Y. (2019). Reviews and price on online platforms: Evidence from sentiment analysis of Airbnb reviews in Boston. *Regional Science and Urban Economics, 75*, 22-34.

Luo, Y., & Tang, R. (2019). Understanding hidden dimensions in textual reviews on Airbnb: An application of modified latent aspect rating analysis (LARA). *International Journal of Hospitality Management, 80*, 144-154.

Qiang, J., Qian, Z., Li, Y., Yuan, Y., & Wu, X. (2020). Short Text Topic Modeling Techniques, Applications, and Performance: A Survey. *IEEE Transactions on Knowledge and Data Engineering*, 1-1.

Zhang, J. (2019). What's yours is mine: exploring customer voice on Airbnb using text-mining approaches. *Journal of Consumer Marketing, 36*, 655-665.